# A REVIEW OF THE ETHICS OF ARTIFICIAL INTELLIGENCE AND ITS APPLICATIONS IN THE UNITED STATES

Taiwo Esther, Akinsola, Ahmed Tella, Edward, Makinde, Kolade, Akinwande, Mayowa

Department of Computer Science, Austin Peay State University, Clarksville USA.

## ABSTRACT

*This study is focused on the ethics of Artificial Intelligence and its application in the United States, the paper highlights the impact AI has in every sector of the US economy and multiple facets of the technological space and the resultant effect on entities spanning businesses, government, academia, and civil society. There is a need for ethical considerations as these entities are beginning to depend on AI for delivering various crucial tasks, which immensely influence their operations, decision-making, and interactions with each other. The adoption of ethical principles, guidelines, and standards of work is therefore required throughout the entire process of AI development, deployment, and usage to ensure responsible and ethical AI practices. Our discussion explores eleven fundamental 'ethical principles' structured as overarching themes. These encompass Transparency, Justice, Fairness, Equity, Non-Maleficence, Responsibility, Accountability, Privacy, Beneficence, Freedom, Autonomy, Trust, Dignity, Sustainability, and Solidarity. These principles collectively serve as a guiding framework, directing the ethical path for the responsible development, deployment, and utilization of artificial intelligence (AI) technologies across diverse sectors and entities within the United States. The paper also discusses the revolutionary impact of AI applications, such as Machine Learning, and explores various approaches used to implement AI ethics. This examination is crucial to address the growing concerns surrounding the inherent risks associated with the widespread use of artificial intelligence.*

## 1. INTRODUCTION

Since the advent of artificial intelligence, various applications have been developed that have assisted human productivity and alleviated human effort, resulting in efficient time management. Artificial intelligence has aided businesses, healthcare, information technology, banking, transportation, and robots. The term "artificial intelligence" refers to reproducing human intelligence processes using machines, specifically computer systems[1].Artificial intelligence allows the United States of America to run more efficiently, produce cleaner products, reduce adverse environmental impacts, promote public safety, and improve human health. Until recently, conversations around "AI ethics" were limited to academic institutions and non-profit organizations. AI could be dangerous to humans and corporations if not employed ethically. Currently, the world's largest technology companies — Microsoft, Facebook, Twitter, Google, and a few other Fortune 500 companies — are forming fast-growing teams to address the ethical issues that arise from the widespread collection, analysis, and use of massive amounts of data, particularly data used to train machine learning models, or artificial intelligence. Most businesses struggle with data and AI ethics through ad hoc talks on a per-product basis, despite the costs of getting it wrong. Teams either miss risks, rush to address problems as they arise, or cross their fingers in the hope that the issue will go away when there is no clear methodology in place for how to detect, evaluate, and mitigate the risks. Companies that have sought to solve the issue at scale have tended to create tight, imprecise, and overly broad procedures, resulting in false positives in risk detection and hindered output. These issues multiply by an order of magnitude when third-party providers are introduced, who may or may not be thinking about these issues at all [2]. Companies must devise a risk-mitigation strategy that addresses how to exploit data and develop AI products while avoiding ethical problems. To operationalize data and AI ethics, it is necessary to identify ethical hazards throughout the company, from IT to HR to marketing to product development.

AI ethics establishes principles, guidelines, and standards governing the responsible development, deployment, and utilization of artificial intelligence (AI) technologies. It encompasses considerations related to morality, societal impact, and legal implications associated with AI systems and their effects on individuals, communities, and the environment. Ethical concerns regarding AI have a historical foundation in the early stages of digital computing[3]. Nevertheless, these concerns have garnered significant attention and relevance in recent years, primarily due to the rapid advancements in AI technology. Addressing AI ethics necessitates the creation of protocols, regulations, and best practices to ensure that AI systems are crafted and utilized in a manner that prioritizes equity, transparency, accountability, and the welfare of individuals and society at large. Achieving this objective entails a collaborative, interdisciplinary approach that involves technologists, ethicists, policymakers, and stakeholders in striking a harmonious balance between technological progress and ethical considerations.

## 2. LITERATURE REVIEW

### 2.1. AI Development

Artificial intelligence (AI) systems are different from expert systems - "collections of rules programmed by humans in the form of if-then statements——are not part of AI since they cannot learn autonomously from external data. Expert systems represent a different approach altogether since they assume that human intelligence can be formalized through rules and hence reconstructed in a top-down approach (also called symbolic or knowledge-based approach)" [4]. Expert systems perform poorly in comparison to AI systems in tasks such as vision, sound, and language understanding. AI systems are only as smart as the data used to train them since they are essentially nothing more than fancy curve-fitting machines.According to [5] in is publication classified AI systems into three groups Analytical AI, Human Inspired AI, and Humanized AI. This classification is based on the intelligence capabilities exhibited by the AI systems (cognitive, emotional, social intelligence, and Artistic creativity)

**Types of AI systems**

| | Expert Systems | Analytical AI | Human-Inspired AI | Humanized AI | Human Beings |
|---|---|---|---|---|---|
| Cognitive Intelligence | ✗ | ✓ | ✓ | ✓ | ✓ |
| Emotional Intelligence | ✗ | ✗ | ✓ | ✓ | ✓ |
| Social Intelligence | ✗ | ✗ | ✗ | ✓ | ✓ |
| Artistic Creativity | ✗ | ✗ | ✗ | ✗ | ✓ |
| | | Supervised Learning, Unsupervised Learning, Reinforcement Learning | | | |

`

Figure 1. Type of AI systems [4]

The benchmark set for AI is the Turing test where if a human is interacting with another human and a machine and is unable to distinguish the machine from the human, then the machine is said to be intelligent.

However, just as humans can never truly understand how chimpanzees think, even though they share 99% of our DNA, we will not be able to understand how an AI system thinks. This limits our ability to control such systems, which again makes them appear more dangerous than useful[5].

## 2.2. Application of AI

In an era characterized by exceptional technological advancement, the use of artificial intelligence (AI) exemplifies remarkable versatility, extending its reach across a vast expanse of industries and sectors. AI is emerging as a vital and powerful tool for addressing difficult challenges in critical domains such as healthcare, education, security, retail, manufacturing, transportation, and data analytics.

AI's ever-evolving capabilities have ushered in transformative solutions, revolutionizing traditional paradigms across these sectors. Its profound impact is evident in the augmentation of diagnostic precision in healthcare, the enhancement of personalized learning experiences in education, the reinforcement of security measures, the optimization of retail operations, the streamlining of manufacturing processes, the evolution of transportation systems, and the extraction of valuable insights from vast datasets.

As AI continues to advance and adapt, it not only reshapes industries but also charts a course toward innovative possibilities. This paradigm shift underscores the imperative for stakeholders to fully embrace and harness AI's potential, for it is through the strategic integration of AI that we navigate the complex challenges and seize the abundant opportunities that define our modern world.

### 2.2.1. AI in Healthcare

We have witnessed the successful application of AI in various healthcare domains, including precision treatment, AI-driven drug discovery, clinical trials, and patient care [2]. Within healthcare, one of the most prevalent applications of machine learning is precision medicine. Precision medicine involves predicting the most effective treatment protocols for individual patients, a determination made based on the patient's historical data [5]. This predictive process relies on training the model using datasets, a method referred to as supervised learning. This type of predictive modeling uses past patient data to uncover patterns and correlations between patient features and treatment outcomes. Machine learning algorithms can detect minor trends and anticipate the best treatment options for new patients by evaluating massive volumes of patient data. The ability to personalize treatments to the individual needs of each patient holds enormous potential for improving healthcare outcomes in the context of precision medicine. It enables healthcare practitioners to shift away from one-size-fits-all approaches and give tailored care that increases the likelihood of successful treatment while decreasing potential negative effects.

Furthermore, AI has proven its utility in drug discovery by speeding up the identification of prospective drug candidates and streamlining the drug development process. AI systems can sift through massive chemical and biological information to discover interesting compounds for future investigation, thereby speeding up the delivery of new medicines to patients. AI techniques are being used to classify candidate compounds in terms of their activity and other properties. For example, a predictive model capable of approximating the strength of binding for candidate molecules may be developed based on experimentally measured binding affinities for a range of substrates. Such predictors can subsequently be used for fast in silico screening and identification of potential drugs with desired properties" [6]. The emergence of QSAR (Quantitative Structure-Activity Relationships) methodologies marked a significant departure from traditional trial-and-error drug discovery processes, offering a more systematic and data-driven approach. By leveraging computational models and datasets, QSAR allows researchers to make informed predictions about the biological activity and safety profiles of novel compounds, thus saving time and resources in the drug development pipeline.

QSAR-based models encounter several challenges, including limitations arising from small training datasets, errors in experimental data within these datasets, and the absence of rigorous experimental validations. To address these hurdles, more recently developed AI methodologies, such as Deep Learning (DL) and related modeling techniques, have gained prominence for the safety and efficacy assessment of drug molecules. These approaches leverage extensive datasets and advanced analytical tools for drug discovery and evaluation. An illustrative example of this transition can be traced back to 2012 when Merck sponsored a QSAR Machine Learning (ML) challenge. This initiative aimed to assess the potential advantages of DL in the pharmaceutical industry's drug discovery process. The results were striking, as DL models exhibited notably higher predictive accuracy when compared to traditional ML approaches. This improvement was particularly evident across 15 absorption, distribution, metabolism, excretion, and toxicity (ADMET) datasets encompassing various drug candidates [7]. The adoption of DL and related AI methodologies signifies a paradigm shift in the field of drug discovery. These data-intensive approaches offer the promise of enhanced predictive capabilities and more robust models, ultimately facilitating the identification of novel drug candidates with improved safety and efficacy profiles. As the pharmaceutical industry continues to embrace AI-driven methodologies, the potential for groundbreaking advancements in drug development remains both exciting and transformative.

In clinical trials, the synergy between AI and healthcare plays a pivotal role in streamlining and expediting patient recruitment, data analysis and monitoring, thus eliminating time-consuming procedures. AI-powered clinical trials exhibit remarkable capabilities in handling vast datasets and delivering highly precise and reliable outcomes [4]. Two significant factors contributing to the failure of clinical trials pre-AI are challenges related to patient cohort selection and recruitment processes, particularly in bringing the most suitable patients to trials promptly. Additionally, the absence of a robust technical infrastructure to manage the complexities of conducting trials, especially in later phases, hampers reliable adherence control, patient monitoring, and clinical endpoint detection systems. Artificial Intelligence (AI) has the potential to address these limitations inherent in current clinical trial design.

Machine learning (ML), specifically focusing on deep learning (DL), can autonomously discern meaningful patterns within vast datasets encompassing text, speech, or images. Natural language processing (NLP) enables the comprehension and correlation of content in written or spoken language, while human-machine interfaces (HMIs), facilitate natural information exchange between computers and humans. These AI capabilities can be harnessed for various purposes within clinical trial design [8]. AI is pivotal in-patient recruitment, data analysis, and monitoring. By automating many aspects of trial management, AI can reduce costs, speed up the research process, and enhance the data processing of large amounts of data. Patient care, too, has seen significant advancements through AI. Machine learning algorithms can assist healthcare professionals in diagnosing diseases, predicting patient outcomes, and optimizing treatment plans. Additionally, AI-driven tools can help manage patient records more efficiently, reducing administrative burdens and allowing clinicians to focus more on patient care.

The progression of AI in healthcare is rapidly advancing, with ongoing research and efforts aimed at addressing AI's limitations and developing innovative solutions to enhance healthcare. While we may be enthusiastic about the potential benefits of these solutions, it is essential to exercise caution. Establishing and vigilantly monitoring ethical standards and principles is imperative to ensure the responsible deployment of AI. This ensures that the advantages of AI in healthcare are realized while minimizing potential risks. The primary ethical concerns arising from AI's use in healthcare encompass liability in cases of medical errors, healthcare professionals' comprehension of machine-learning prediction generation, patients' understanding and control over the use of machine-learning tools in their healthcare, as well as issues related to biases, privacy, security, and the management of patient data[9].It is essential to prioritize ethical training and adaptability in AI systems. Implementing best practices to counteract bias introduced through training data is crucial in preventing data-driven AI software from perpetuating or worsening pre-existing clinical biases. Additionally, developers must evaluate how the data inputs essential for their software may impact the scalability of their products when applied to diverse settings distinct from the original data source used for algorithm training [10]. Lastly, it is imperative to establish and potentially evolve best practices and paradigms to safeguard patient privacy effectively.

### 2.2.2. AI in Technology

In the quest for achieving human flourishing through technology development, there's a growing emphasis on fostering better practices and behaviors. The development of AI technologies and this improvement extends to the cultivation of ethical values that contribute to creating "technology for good." The term "technology for good" has gained popularity across research, industry, and policy circles, each operating at different scales and with varying priorities [11]. Previous research in this domain has explored concepts like information technology for the betterment of society, technology for improving governance, and tech with a positive social impact.

AI is making significant inroads in various industries, providing valuable services, and enhancing capabilities [12]:

1. Virtual Assistants: Many industries are adopting virtual assistants to aid users, exemplified by Tesla's Tesla Bot, which offers real-time assistance.
2. Self-Driving Cars: Organizations are developing self-driving cars to improve safety and security during journeys compared to traditional manual driving.
3. AI in Robotics: AI is revolutionizing robotics by enabling robots to draw upon past experiences to solve tasks, expanding their capabilities beyond repetitive functions.
4. Humanoid Robots: AI algorithms are being deployed in humanoid robots like Sophia and Erica, allowing them to mimic human behavior and communication, blurring the lines between robots and humans.

### 2.2.3. AI in Military

The Department of Defense (DOD) is establishing AI ethical principles that build upon its existing ethical framework derived from the U.S. Constitution, Title 10 of the U.S. Code, the Law of War, international treaties, and established norms. These principles are designed to address the unique ethical challenges of using AI in the military. They will apply to both combat and non-combat functions and encompass five key areas [13]:

1. Responsibility: DOD personnel will exercise care and accountability in developing, deploying, and using AI capabilities.

2. Fairness: Measures will be taken to minimize unintentional bias in AI capabilities.
3. Transparency: AI capabilities will be developed with transparency and understanding, utilizing auditable methodologies and precise documentation.
4. Reliability: AI capabilities will be rigorously tested and assured for safety, security, and effectiveness within defined purposes.
5. Governance: AI capabilities will be designed to detect and prevent unintended consequences and have mechanisms to disengage or deactivate systems displaying unintended behavior.

### 2.2.4. AI in Financial Services

There is often an argument in finance that regulatory measures have effectively addressed systemic risks, rendering ethical concerns about these risks unnecessary. Can legal regulations substitute ethical considerations? Certainly, systemic risks have been a primary focus of financial sector regulators over the past decade, as evidenced by initiatives such as Basel III, MiFID II, the German Banking Act, and the Regulation Systems Compliance and Integrity implemented by the U.S. Securities and Exchange Commission (SEC). Simultaneously, these regulatory efforts impose legal responsibilities on individuals and entities designing and implementing trading algorithms. Regulators often adopt a simplifying approach, assuming a clear, cause-and-effect relationship where actions at the individual level (e.g., actions by investment firms) directly influence undesirable outcomes at the macro level (e.g., financial market instability). For instance, regulatory authorities mandate that investment firms adhere to specified order limits to prevent erroneous orders that could disrupt the market. They also require these firms to continuously test and monitor their algorithms, implement circuit breakers for emergency trading halts and regularly provide self-assessment and validation reports on risk mitigation measures [14].

A proposed ethical intermediary should collaborate with established regulatory and oversight bodies that focus on addressing systemic risks within financial markets. These bodies, including institutions like the Bank for International Settlements and the Financial Stability Board, have already amassed significant knowledge in this area. Furthermore, the entity responsible for mediating between AI-driven markets and ethical concerns should actively seek input and ethical considerations from two main stakeholders: companies utilizing AI and their employees on one side and the public and citizens on the other side. Insights gathered from these consultations could enhance the accountability framework and later be incorporated into ethical codes and guidelines [15].

### 2.2.5. AI in Education

Within the realm of education, the integration of artificial intelligence (AI) unfolds through three distinct paradigms: AI-guided - with learners taking on the role of active participants; AI-assisted - fostering collaboration between learners and technology; and AI-empowered, where learners use AI for self-directed exploration, research, and critical thinking in their educational journey [16]. Artificial Intelligence in Education (AIEd) has demonstrated its potential to benefit not only learners but also educators. A noteworthy manifestation of this potential is the development of automated assessment systems designed to assist teachers in evaluating students' knowledge. Additionally, facial recognition systems have emerged as valuable tools providing insights into learners' behaviors[17].Like many other sectors where AI adoption has taken place, the education sector has not been immune to ethical concerns from various stakeholders. These span a broad spectrum of critical issues, including privacy, which revolves around ensuring compliance with regulations such as the Family Educational Rights and Privacy Act (FERPA) concerning the collection and utilization of data to train AI models. fairness, accountability, transparency, bias, autonomy, agency, and inclusion.

In a broader context, it becomes paramount to distinguishbetween merely adhering to ethical standards and conducting activities in an ethically sound manner. This distinction underscores the need for a profound comprehension of pedagogical choices that align with ethical principles. Furthermore, it emphasizes the significance of a steadfast commitment to addressing the potential unintended consequences that invariably accompany the deployment of AI in educational initiatives. [Ethics of AI in Education: Towards a Community-Wide Framework [18].

# 3. AI ETHICAL PRINCIPLES

In the AI ethics literature, numerous terms and expressions have surfaced. For instance, in a thorough review conducted in 2019 on AI ethics guidelines, eleven key 'ethical principles' were identified. These include Transparency, Justice, Fairness, Equity, Non-Maleficence, Responsibility, Accountability, Privacy, Beneficence, Freedom, Autonomy, Trust, Dignity, Sustainability, and Solidarity [19].

As mentioned earlier, there is some overlap among these terms, and they often require clarification. In this overview, we leverage the growing engineering expertise that intersects with ethical principles. Specifically, in the following section, we identify and explore seven overarching themes that aim to bridge the gap between the need to incorporate ethics into engineering and systems. These principles encompass human agency and oversight, safety, privacy, transparency, fairness, and accountability.

Drawing inspiration from Deloitte's Trustworthy AI™ framework, which comprises attributes such as transparency and explainability, fairness and impartiality, robustness and reliability, privacy, safety and security, and responsibility and accountability, these themes can collectively be seen as falling under the broader concept of 'Trustworthy AI' [20].

## 3.1. Transparency and Explainability

This ethical principle in AI underscores the importance of being open and clear, which is critical for building trust and ensuring accountability. Transparency involves disclosing the decisions made regarding the use of AI systems and the methods these systems employ to arrive at their decisions, making the process open to scrutiny and tracking. Various sources recommend sharing more information to improve transparency in AI development and implementation. However, there is considerable variation in the specifics of what should be disclosed. This encompasses details like how AI is used, its code, how it handles data, the evidence supporting its use, any limitations, applicable laws, who is responsible for AI systems, the investments made in AI, and the potential consequences.

Furthermore, it is encouraged that this information is presented in a way that non-experts can understand to ensure accessibility. Additionally, it should be designed to be comprehensible by humans so they can audit it themselves. Data protection agencies and non-profit organizations often suggest audits and making the information auditable, while the private sector more frequently proposes technical solutions.Alternative approaches to promote transparency include oversight, engagement, and mediation with stakeholders and the public. Moreover, enabling individuals to report wrongdoing is another method to foster transparency and accountability in the development and use of AI [19].Information transparency is considered ethically beneficial when it supplies the necessary information to support ethical principles, either by depending on them or by explaining how information is restricted through regulation. Conversely, ethical principles can be compromised when incorrect information (misinformation) is shared or when insufficient or excessive information is disclosed[21].

## 3.2. Fairness and Impartiality

This concept relates to ensuring that AI systems treat individuals and groups fairly and without bias. When appropriately calibrated, AI can aid humans in making more impartial decisions, counteracting human prejudices, and fostering inclusivity. A key question is which definition(s) of fairness or justice to adopt, given that there are various, sometimes conflicting, theories of fairness, such as corrective, distributive, procedural, substantive, comparative, and others. Additionally, there's a consideration of the context in which discussions of fairness and justice occur, whether within the scope of political communities (concerning citizenship rights) or universal human issues [22]. If applicable, there's also the challenge of defining demographic factors like gender, nationality, race, socio-economic background, etc.

1. Unbiasedness:Emphasizing the concern regarding biases present in datasets underscores the critical importance of obtaining and managing accurate, inclusive, and diverse data, particularly when it comes to training data [16]. Problematically, a substantial portion of research on fairness in AI is driven by situations where our foundational data is flawed, reflecting existing discriminatory patterns. Defining various types of data bias, identifying them, and making valid inferences from such data continue to pose significant challenges. Even in cases were providing preferential treatment to disadvantaged groups is deemed an acceptable solution, questions remain about the methods, extent, and timing of such actions. While in some instances, providing preferential treatment might be viewed as a correction for past discrimination stemming from unaccounted variables, in other contexts, it could be interpreted as a form of discrimination itself [23].
2. Accessible:the importance of fair access to AI benefits, particularly regarding public sector involvement, draws attention to AI's effects on the job market and the imperative to address democratic and societal issues.
3. Inclusive:This pertains to the assurance that AI systems are created and used in a manner that is accessible and comprehensible to a diverse audience. The goal is to promote wider adoption of AI technology in society. Developers of AI should convey information about their systems in a manner that is easily comprehensible to the public. This involves using straightforward language and avoiding technical terminology, simplifying users' engagement with AI technology.

## 3.3. Robustness and Reliability

Ensuring that artificial intelligence (AI) systems are robust, dependable, and consistently ethical is essential. This entails designing AI systems that can withstand challenges, operate reliably, gain trust, and consistently adhere to ethical principles. The goal of robustness can be seen as the aim for an AI system to function reliably and accurately even in challenging scenarios. These scenarios might include adversarial interference, errors by implementers, or skewed goal execution in automated learning, especially in reinforcement learning applications. The measure of robustness, therefore, gauges the system's resilience and its ability to function soundly when confronted with difficult conditions, adversarial attacks, disruptions, data tampering, and unwanted behaviors in reinforcement learning [24].Reliability pertains to the system's alignment with its intended purpose and deployment, whereas reproducibility involves consistent performance under identical inputs and conditions. In the context of robustness, this is of utmost importance because an unreliable system that cannot reproduce results erodes trust in the system.

### 3.4. Privacy

Privacy is indispensable for exercising a range of human rights, such as freedom of expression and association, as well as being fundamental personal autonomy, freedom of choice, and broader societal norms [25].

1. Data exploitation: Consumer goods, from smart home gadgets to connected vehicles and mobile apps, are frequently designed to extract data. Consumers often encounter an information imbalance regarding the types and quantities of data that their devices, networks, and platforms generate, handle, or share. As we introduce a growing number of smart and interconnected devices into our residences, workplaces, public areas, and even our bodies, educating the public about this data exploitation becomes more urgent.

2. Data Minimization: Data minimization in privacy and data protection involves using only the required data for a specific purpose. This principle consists of three dimensions: adequacy (data should be sufficient for the purpose), relevance (data should have a justifiable link to the purpose), and necessity (data should be limited to what is required and, when appropriate, deleted when it is no longer needed for the specified purpose).

3. Identification and tracking: AI applications can identify and track individuals across various devices and settings, including their homes, workplaces, and public areas. For instance, even when personal data is typically anonymized within datasets, AI can reverse this process (de-anonymization), blurring the line between personal and non-personal data, which forms the basis of current data protection regulations. Facial recognition technology is another method for tracking and identifying individuals, potentially changing the concept of anonymity in public spaces. Machine learning systems have demonstrated the ability to identify approximately 69% of protesters who used caps and scarves to conceal their faces[26].

### 3.5. Accountability

Humans create algorithms and the data they rely on, and there is always a human ultimately responsible for the decisions influenced by an algorithm. Blaming "the algorithm" is not an acceptable excuse when algorithmic systems make errors or lead to unintended consequences, even in cases involving machine learning processes. Since such algorithm-driven decisions can have significant societal consequences, this document aims to guide developers and product managers in designing and implementing algorithmic systems in a publicly accountable way. In this context, accountability entails reporting, explaining, or justifying algorithmic decision-making and addressing and mitigating any adverse social impacts or potential harms[27].

The development of systems, the decision-making processes, the logic behind those decisions, the assignment of responsibilities for decision-makers, and the assessment of impacts, risks, and harms all come under the umbrella of accountability. As mentioned earlier, accountability is about understanding who made the decisions, their decision-making methods, and the systems or tools to measure and monitor these decisions, essentially encompassing governance.

### 3.6. Responsibility

The ethical principle guiding AI can be traced back to humans. The guidelines emphasize that humans are responsible for decisions influenced by AI. Human decisions should be well-informed and not forced into critical domains affecting individuals. Humans must constantly make decisions that involve taking human life, and those who approve AI for illegal purposes are accountable. In situations where AI causes harm, if it's used as intended and proven reliable, responsibility should not be attributed to those who were involved in its development or usage [28].

## 3.7. Safety and Security

This ethical principle focuses on preventing harm, which is defined as adverse impacts on human well-being, encompassing psychological, social, and environmental aspects of well-being. The safety risks encountered in your AI project will depend on various factors, including the algorithms and machine learning techniques employed in the intended applications, the source of your data, how you define your objectives, and the problem domain you're addressing. Regardless of these variables, it's considered a best practice to incorporate safety considerations related to accuracy, reliability, security, and robustness at every stage of your AI project's life cycle. This should involve the rigorous application of testing, validation, verification, and ongoing system safety monitoring [24]. Additionally, self-assessments for AI safety should be conducted at each workflow stage to evaluate how your design and implementation practices align with safety objectives. These self-assessments should be continuously documented and made reviewable.

## 4. IMPLEMENTING THE PRACTICAL USE OF ETHICAL AI APPLICATIONS

AI predominantly relies on the utilization of Machine Learning (ML) algorithms. Significant attention has been given to ethical and bias-related challenges within the domain of AI applications, particularly concerning ML algorithms. The biases held by team members responsible for selecting datasets used to train these algorithms can have a notable impact on the outcomes generated by the algorithms, which, in turn, raises ethical concerns regarding the results they produce. When confronted with new information, subjecting the algorithm's findings to rigorous scrutiny is imperative. This entails conducting a comprehensive assessment of the data utilized for training the algorithms, aiming to identify and rectify any biases present in the dataset. Additionally, it is crucial to ensure that the appropriate algorithms are applied to specific tasks. Failing to provide adequate oversight can result in these algorithms perpetuating, amplifying, and compounding biases in their outcomes and their interpretation and knowledge.

Individuals' beliefs and values regarding AI implementation are pivotal in shaping their perspective on ethical AI usage, particularly within machine learning. Detecting and mitigating potential biases that could impact research outcomes is paramount in any research endeavor. This principle applies equally to the data selection process and the development and training of machine learning algorithms, necessitating identifying, mitigating, and eliminating biases. Bias can manifest in various aspects of our decision-making, including selecting data and emphasizing specific data types. Recognizing the presence of bias and taking proactive measures to address it is essential to ensure that bias does not influence the design, development, and execution of AI applications [29]. Forming diverse teams that bring a wide array of ideas, experiences, and knowledge to the forefront plays a pivotal role in addressing bias in AI and enhancing its ethical standards.

Beginning with a well-structured AI Policy and Standards framework is essential to implement AI Ethics effectively. The following steps provide a guide for translating ethical AI principles into practical applications:

1. Embrace and adhere to AI standards that incorporate criteria for evaluating the ethical aspects of AI applications and defining measures for eliminating bias.
2. Form a Diverse AI Product Development Team:

- Promote collaboration, knowledge sharing, and utilizing diverse perspectives, experiences, and cultural backgrounds to foster various ideas.
- Diversity in thought serves as a catalyst for innovation, enabling organizations to create unique and improved AI products and driving progress and growth.

3. Assemble a Diverse Team for AI Application Design, Development, and Implementation:
   - A diverse team brings a variety of perspectives, especially during the data selection and cleansing processes for AI applications utilizing Machine Learning.

4. Develop AI Applications with a Human-Centered Approach:
   - Prioritize the inclusivity and well-being of individuals served by AI applications, adhering to human-centric values, and promoting fairness.
   - Design, develop, and implement human-centered AI applications with transparency, robustness, safety, and accountability for their outcomes and decisions influenced by AI.

5. Establish Key Performance Indicators (KPIs) and Metrics for AI:
   - To advance the implementation of AI standards, guidelines, and principles, introduce standardized metrics for assessing AI systems consistently.
   - Develop evidence-based metrics and KPIs to evaluate deployed AI applications' performance continuously.

By following these steps, organizations can effectively integrate ethical AI principles into their practices and ensure responsible AI development and deployment.

In recent years, companies have increasingly harnessed the power of data and artificial intelligence to create scalable solutions. However, this rapid adoption of AI has brought with it significant reputational, regulatory, and legal risks. For example, Los Angeles has taken legal action against IBM for allegedly misappropriating data collected through its widely used weather app. Optum faced regulatory investigations for an algorithm that reportedly favored white patients over sicker black patients. Similarly, Goldman Sachs came under regulatory scrutiny for an AI algorithm accused of discriminating against women in credit limit determinations on their Apple cards. Notably, Facebook faced a major backlash when it granted Cambridge Analytica access to personal data from over 50 million users [2].

This shift towards addressing data and AI ethics is not limited to nonprofits and academia; even the world's largest tech companies, including Microsoft, Facebook, Twitter, and Google, have established dedicated teams to tackle ethical challenges arising from extensive data collection and AI model training. They have recognized that failing to operationalize data and AI ethics poses a significant threat to their bottom line.Companies that neglect data and AI ethics risk not only reputational damage and regulatory penalties but also wasted resources and inefficiencies in product development. An illustrative example is Amazon, which reportedly spent years developing AI hiring software but eventually abandoned the project due to its inability to create a model free from systematic gender bias. Similarly, Sidewalk Labs, a subsidiary of Google, faced substantial opposition and eventually scrapped a 'smart city' project in Toronto due to a lack of clear ethical standards, resulting in a loss of significant time and financial resources.

However, most companies still struggle with data and AI ethics on an ad-hoc, per-product basis. The absence of clear protocols for identifying, evaluating, and mitigating ethical risks can lead to oversights or reactive problem-solving. Scaling up efforts to address these issues often results in overly broad policies that generate false positives and impede production, especially when third-party vendors are involved.Considering these challenges, companies require a comprehensive plan for mitigating ethical risks in data and AI. An operationalized approach must systematically identify ethical risks across various departments, from IT to HR to marketing and product development.

## 4.1. Two Practical Approaches for Implementing AI Ethics

In addition to the insights above, two significant practical approaches for implementing ethics in AI systems are worth noting. The first approach is exemplified by the IEEE Global Initiative on Ethics of Autonomous and Intelligent Systems ("The IEEE Global Initiative"), while the second approach is driven by the World Economic Forum's project on Artificial Intelligence and Machine Learning.

### The IEEE Global Initiative

The IEEE Global Initiative, initiated in December 2015, strives to ensure that technologists involved in AI development prioritize ethical considerations. A central document of this initiative, titled "Ethically Aligned Design – A Vision for Prioritizing Human Well-being with Autonomous and Intelligent Systems," outlines concrete operational standards under the IEEE P7000™ series. These standards offer guidance for designers working on AI and autonomous systems. Key proposals within this series address transparency, data privacy, algorithmic bias, child and student data governance, and ethical considerations in robotics and automation systems [30].

### World Economic Forum's Project on Artificial Intelligence and Machine Learning

The World Economic Forum is crucial in fostering international public-private collaborations in AI governance. They emphasize ethics, social inclusion, and human-centered design in their projects. The Forum is establishing Centers for the Fourth Industrial Revolution in key cities worldwide, working closely with governments, businesses, academics, and civil society. Some ongoing projects include:

1. **Unlocking Public Sector AI:** This project focuses on establishing baseline standards for ethical procurement and deployment of AI in public institutions. It addresses concerns related to bias, privacy, accountability, and transparency.
2. **AI Board Leadership Toolkit:** As AI becomes integral to various industries, this toolkit aids corporate leaders in understanding the benefits and ethical challenges of AI adoption. It offers practical tools for responsible decision-making, including the appointment of Chief Values Officers and Ethics Advisory Boards.
3. **Generation AI:** This project centers on creating standards to protect children in the age of AI. AI is increasingly integrated into children's products and education tools. Guidelines aim to safeguard privacy, involve parents in understanding AI designs, and prevent biases in AI training data.
4. **Teaching AI Ethics**: Recognizing the need for AI engineers to understand ethical implications, this initiative provides educational resources for integrating ethical considerations into technical curricula.
5. **The Regulator of the Future**: To ensure effective AI governance, this project reimagines the role of regulators, facilitating a collaborative approach involving citizens, companies, and governments in understanding and regulating advanced technologies.

## 4.2. Limitations

The rapidly changing AI landscape requires continuous adaptation and refinement of ethical principles and guidelines to remain in sync with technological advancements. It's crucial to acknowledge that what constitutes ethical conduct may shift as AI capabilities and applications continue to expand. Given the multifaceted nature of AI ethics, a more thorough examination is necessary, tailored to specific industries and contexts. While the study provides a comprehensive overview of essential ethical principles and practical strategies for responsible AI development and deployment, addressing the evolving complexities and persistent challenges in AI ethics necessitates in-depth, specialized investigations conducted by researchers, policymakers, and practitioners within their respective fields.

# 5. CONCLUSION AND RECOMMENDATIONS

Our exploration into the ethics of artificial intelligence and its broad applications in the United States has unveiled a dynamic and intricate landscape. The United States holds a leading position in AI innovation, where the potential of transformative technologies is balanced with the ethical responsibilities that come with them. As we wrap up this journey, it's crucial to reflect on the distinctive ethical challenges and opportunities that the U.S. encounters in its AI advancement. AI's versatile applications, spanning healthcare, technology, finance, and the military, are already reshaping core sectors of the U.S. economy and society. The potential of AI is evident in its capacity to boost productivity, enhance healthcare outcomes, stimulate economic growth, and strengthen national security. This technology has the potential to revolutionize how businesses function, how citizens access services, and how policymakers tackle societal issues.

However, the tremendous potential of AI brings forth profound ethical considerations. Throughout our exploration, we have emphasized the critical need to address these ethical dimensions. The United States, being a leader in AI development and deployment, bears a unique responsibility in shaping the ethical path of this technology. Transparency remains a foundational principle of ethical AI in the United. States. Citizens and stakeholders must have confidence that AI systems are designed and deployed with transparency, offering clear explanations of their operations. Transparency fosters trust, which is fundamental for the acceptance of AI by society. The principle of fairness and equity holds immense importance in a nation that values equal opportunity. Ensuring that AI systems do not perpetuate biases or discriminate against specific groups is a fundamental ethical imperative. The United States must lead in developing fair and unbiased AI technologies. Accountability and responsibility extend beyond individual actors to include organizations, policymakers, and a whole. The U.S. should establish accountability frameworks, clearly defining those responsible for AI decisions and ensuring mechanisms to address adverse consequences, robust privacy, and data protection framework.

Since AI systems process extensive personal data, safeguarding individual privacy rights is ethically mandated. In a world where AI is increasingly integrated into critical systems, ensuring the security and safety of AI technologies is of utmost importance. The U.S. must lead efforts to prevent AI-related security breaches and prioritize the development of secure AI systems.

As we conclude this exploration, it's vital to acknowledge that the ethical boundaries of AI in the United States are dynamic, requiring continuous attention and adaptation. The choices made today will shape the ethical development of AI in the United States and its global impact. In this era of remarkable technological progress, the United States has an opportunity to exhibit ethical leadership in AI. By adhering to ethical principles, advocating transparency, upholding fairness, and embracing accountability, the nation can serve as a guiding beacon for responsible AI development worldwide. Ethics should be at the heart of AI innovation in the United States, enabling the nation to unlock the full potential of AI technologies while managing risks and ensuring equitable distribution of benefits across society. Ethical AI is a shared responsibility involving government, industry, academia, and civil society. Collaboration and dialogue among these stakeholders are imperative for advancing AI ethically. Through collective efforts, the United States can establish a robust ethical framework for AI that sets a global standard. The ethics of artificial intelligence and its application in the United States present challenges and opportunities. As a pioneer in AI, the United States can shape the future of this technology while upholding ethical values. The decisions made today will resonate through generations, defining AI's role in American society and its position on the global stage. Let us, as responsible stewards of AI, navigate these ethical frontiers with wisdom, foresight, and an unwavering commitment to the ethical principles that underpin the future of artificial intelligence in the United States.

## AUTHORS

**Esther Taiwo** a dedicated professional in the IT and data field holds a Bachelor of Science (BSc.) degree in Mathematics and Statistics. Currently, she is pursuing a Master of Science (MSc.) degree in Computer Science and Quantitative Methods. Esther's passion lies in utilizing data to foster innovation and enhance solutions that positively impact our world.

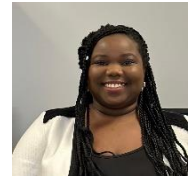

**Akinsola Ahmed** is a graduate student at Austin Peay State University in Clarksville, Tennessee, pursuing a Master of Science in Computer Science. He is passionate about information security and AI, and he believes these technologies have the potential to make the world a safer and more equitable place as such he's on a journey to leave an indelible mark in the world of technology, inspiring the next generation of innovators.

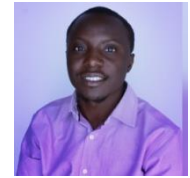

**Edward Tella** is a graduate student of Predictive Analytics at Austin Peay State University, Tennessee and holds a BSc in Systems Engineering from the University of Lagos, Nigeria. Edward has a wealth of professional experience as a Product Manager at two of the largest banks in Africa. His commitment to advancing the field of predictive analytics, coupled with his proven track record of success, makes him poised to drive.

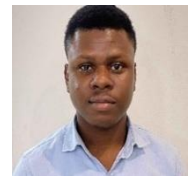

**Kolade Makinde** is a current Master (MSc) student studying Computer science and quantitative Methods with a concentration in Predictive, with an Engineering background, it was less tasking taking up the role of a data analyst in the technology space. He has a keen interest in finance and business management and believes he would be a part of the fast-paced global solution provider in this field.

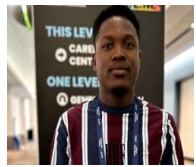

**Mayowa Akinwande** is a results-driven professional with a strong background in operations, IT, and data analysis. He holds a Bachelor of Science (BSc.) degree in Physics and Education, showcasing his commitment to both technical and educational pursuits. Currently, Mayowa is pursuing a Master of Science (MSc.) degree in Computer Science and Quantitative Methods, furthering his expertise in the ever-evolving field of technology and data analytics. He aims to leverage his data-driven skills and insights to drive positive change on a global scale.

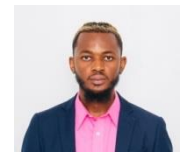